\documentclass[journal]{IEEEtran}

\usepackage{caption2}

\ifCLASSINFOpdf
   \usepackage[pdftex]{graphicx}
   \DeclareGraphicsExtensions{.pdf,.jpeg,.png}
\else
\fi
\usepackage{multirow}

\usepackage[cmex10]{amsmath}

\usepackage[ruled,norelsize]{algorithm2e}

\makeatletter
\newcommand{\removelatexerror}{\let\@latex@error\@gobble}
\makeatother

\usepackage{array}
\ifCLASSOPTIONcompsoc
  \usepackage[caption=false,font=normalsize,labelfont=sf,textfont=sf]{subfig}
\else
  \usepackage[caption=false,font=footnotesize]{subfig}
\fi

\usepackage{color}

\graphicspath{{images/}}

\begin{document}
\title{Towards predicting the likeability of fashion images}

\author{Jinghua Wang, Abrar Abdul Nabi, Gang Wang,~\IEEEmembership{Member,~IEEE,}
Chengde Wan, Tian-Tsong Ng,~\IEEEmembership{Member,~IEEE,} 
% <-this % stops a space
\thanks{Jinghua Wang, Abrar Abdul Nabi, and Gang Wang are with the School of Electrical and Electronic Engineering, Nanyang Technological University, Singapore.

Chengde Wan is with Computer Vision Laboratory, ETH Zentrum, Switzerland.

Tian-Tsong Ng is with Institute for Infocomm Research, Agency for Science, Technology and Research (A*STAR),  Singapore.
}% <-this % stops a space
%\thanks{Manuscript received April 19, 2005; revised September 17, 2014.}
}

\maketitle

\begin{abstract}
In this paper, we propose a method for ranking fashion images to find the ones which might be liked by more people. We collect two new datasets from image sharing websites (Pinterest and Polyvore). We represent fashion images based on attributes: semantic attributes and data-driven attributes. 
To learn semantic attributes from limited training data, we use an algorithm on multi-task convolutional neural networks to share visual knowledge among different semantic attribute categories.
To discover data-driven attributes unsupervisedly, we propose an algorithm to simultaneously discover visual clusters and learn fashion-specific feature representations. Given attributes as representations, we propose to learn a ranking SPN (sum product networks) to rank pairs of fashion images. The proposed ranking SPN can capture the high-order correlations of the attributes. We show the effectiveness of our method on our two newly collected datasets.
\end{abstract}


% \fi
\IEEEpeerreviewmaketitle

\section{Introduction}
%\IEEEPARstart{T}{his} demo file is intended to serve as a ``starter file''
%for IEEE journal papers produced under \LaTeX\ using
%IEEEtran.cls version 1.8a and later.
%% You must have at least 2 lines in the paragraph with the drop letter
%% (should never be an issue)
%I wish you the best of success.
%\hfill September 17, 2014

\IEEEPARstart{O}{nline} shopping increases significantly recently \cite{IncreaseOnlineShopping}. According to the research from Wipro Digital in 2013 and 2014, the percentage of shoppers that make the majority of their purchase online grows from 36\% to 61\% in U.S., and from 45\% to 71\% in U.K. More importantly, half of participants in the survey intend to do more shopping online in the future. 

%By contrast, only 4\% of U.S. and 6\% of U.K. shoppers plan to increase their visits to stores.

Among all of the categories, fashion contributes most in the increase of online shopping. It is revealed that clothing saw the biggest year-on-year increase, going up 20.6\% in June 2014 compared with June 2013 \cite{fashionContributeMost}.

Due to its huge profit potential, fashion analysis is receiving increasing attention these
days. To meet the huge
online clothing shopping needs, it demands  computer vision techniques to process fashion data automatically in this area. Representative examples are  clothing retrieval \cite{Street_to_shop_Yanshuicheng}, parsing \cite{paper_doll_parsing_ICCV2013_berg,parsing_clothing_cvpr2012_berg}, and fashion style prediction \cite{HipsterWarsECCV14}.

%
%\begin{figure}[t]
%\begin{center}
%   \includegraphics[width=0.6\linewidth]{figureDresses_ver4}
%\end{center}
%   \caption{Example of dress images from Pinterest.com (best viewed in color). The images in the first row have many more \textit{likes} than the ones in the second row, i.e. the first three images are liked by many more people. For a pair of images, this paper aims to tell which one is liked by more people.}
%\label{fig:examples_low_hig}
%\end{figure}

\begin{figure}[!t]
 \centering
 \includegraphics[width=2.5in]{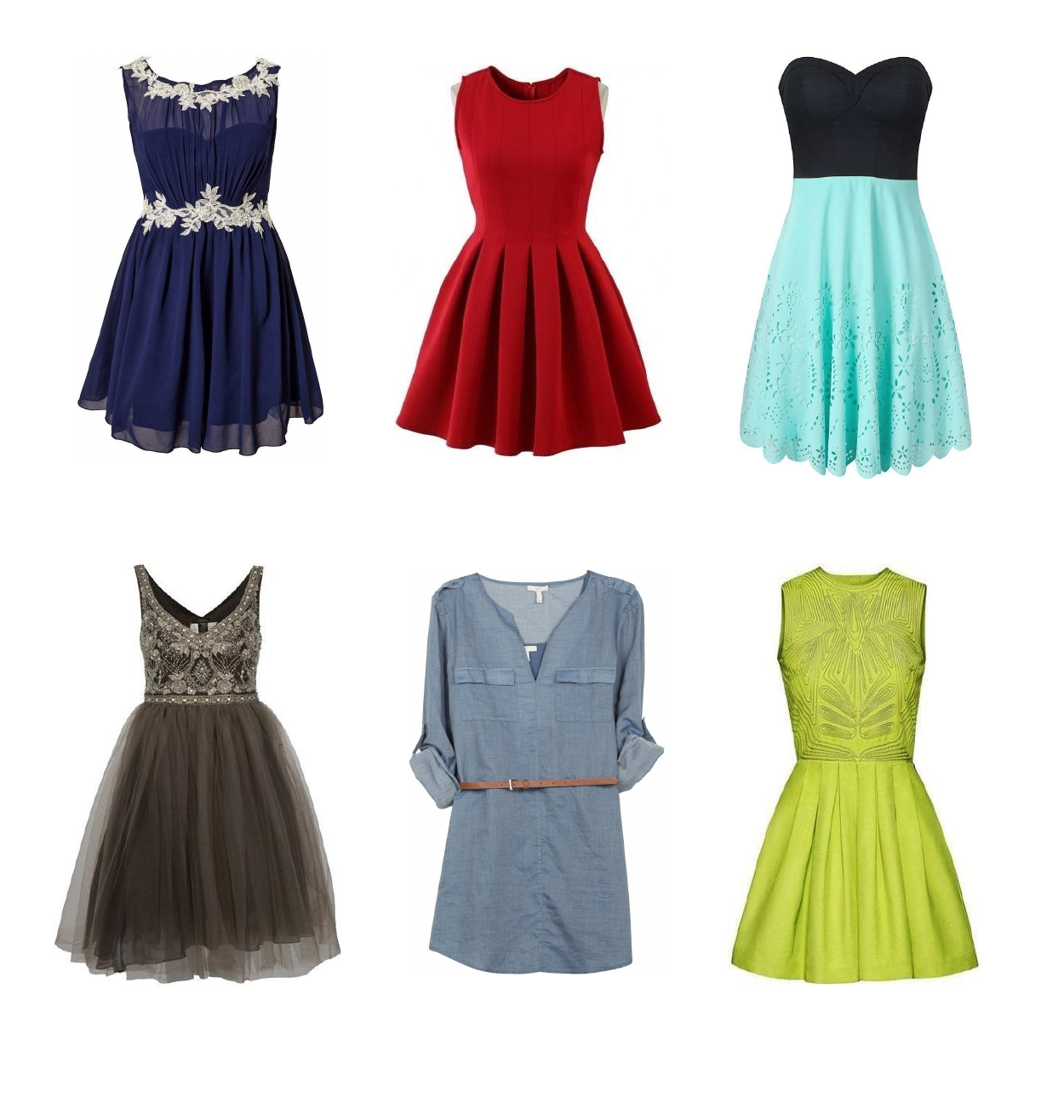}
 \caption{Example of dress images from Pinterest.com (best viewed in color). The images in the first row have many more \textit{likes} than the ones in the second row, i.e. the first three images are liked by many more people. For a pair of images, this paper aims to tell which one is liked by more people.}
 \label{fig:examples_low_hig}
\end{figure}

On fashion shopping and social websites, there is another strong need: determine what fashion images (products) should be put on top (or placed in the first page) to attract the attention of users. Intuitively, such fashion images (products) should be liked by more people. Currently, many websites hire so-called fashion experts to manually select images. But human experts can be biased and they cannot compare between a large amount of images.  To solve this problem, can we build a computerized intelligent program to automatically compare fashion images, and identify those that have potential to be liked by many people? On photo sharing websites such as Pinterest, we can collect millions of people's views on tens of millions of fashion images. From such a large amount of labeled data, we can possibly train a ranking classifier to discover discriminative visual patterns to rank images according to the likeability. People's views can be subjective. In this paper, we focus on learning ranking machines based on data whose rank is agreed by many people. We do not aim at distinguishing fashion images with low consensus.  For example, Figure \ref{fig:examples_low_hig} shows six images from the website (\textit{pinterest.com}). Though few people may have different
opinion, most people would find that the three images in the first
row are more attractive than the ones in the second row.
This is verified by the statistics from the website. Each of
the first three images gains more than 50 \textit{likes}. The last three images gain less than 3 \textit{likes}.

To learn such a ranking machine, we need to represent fashion images in a proper way.
People tend to judge fashion images based on mid-level attributes. In
many cases, people may like a dress because of a particular
semantic attribute such as `major color is red'. Accordingly, in this paper, we resort to representing fashion based on middle-level attributes. However, some attributes are not nameable (in contrast to the nameable semantic attributes). Hence, we also train classifiers to recognize data-driven attributes that represent visual patterns in dress images which don't have names. 

These two types of attributes are learned based on the techniques of deep learning, which has made astonishing progress in many computer vision areas such as object recognition \cite{imageNetCNN}, object detection \cite{object_detection_girshick2013rich}, and OCR \cite{ocr_deeplearning}. To effectively learn the semantic attributes from a limited amount of training data, we use a structure of convolutional neural networks with a shared layer for
multi-task learning, hence the visual knowledge of different attribute classes can be shared to boost the performance of each individual attribute classifier.
Data-driven attribute learning is similar to visual discovery, since we need to identify coherent visual patterns from data in an unsupervised manner. We propose a new algorithm for unsupervised clustering and CNN model learning simultaneously. We first discover visual clusters based on the features of convolutional neural networks, and then learn better convolutional neural networks for our data given the discovered visual clusters, in an iterative manner. Each visual cluster is considered as one data-driven attribute. Using the sliding window method, we know the occurrence of the data-driven attributes in an image.

After fashion images are represented based on the activation vectors of attributes, we propose to learn a ranking SPN (Sum-Product Networks \cite{Poon_Domingos_SUM_PRODUCT_ORIGINAL}) to rank pairs of images. Compared to the traditional ranking machines such as ranking SVM, a ranking SPN can capture the high-order correlation of attributes for ranking. This is very critical in our application: people may like a dress because of the co-occurrence of two or more fashion attributes, they may also dislike a dress because of the co-occurrence even these attributes are attractive individually.

To learn parameters for the ranking SPN, we propose a method to update the SPN based on its evaluations on a pair of images. We use the root value of the SPN to assess the likeability of an image. For two images, if they have quite different number of \textit{likes}, the proposed method updates the parameters to increase the difference between the root values. Otherwise, the proposed method updates the parameters to decrease the difference between these two evaluations.

We test our methods on two datasets collected from Pinterest and Polyvore respectively. The experiments prove the effectiveness of our methods. Our main contributions of this paper are summarized as follows: 

1) we build two new fashion image datasets (i.e. Pinterest and Polyvore) for predicting the likeability of fashion images;

2) we propose an unsupervised method to simultaneously discover data-driven attributes and perform feature learning for our task;

3) we develop a ranking SPN to rank fashion images by modeling the high-order correlation of attributes.

The rest of this paper is organized as follows: Section \ref{Sec:Relatedwork} presents related work of this paper. Section \ref{Sec:Dataset} describes our two datasets. Section \ref{sec:journalAttributeLearning} shows how we learn the semantic and data-driven attributes. Section \ref{Sec:spatialRankingSPN} describes how we learn the parameter of the ranking SPN. Section \ref{Sec:Experiment} presents the experiments. Section \ref{Sec:Conclusion} concludes this paper.

\section{Related Work}
\label{Sec:Relatedwork}

We develop our representations based on attributes learning. Many works show the importance of attributes as mid-level descriptors in visual recognition tasks \cite{decorrelatingAttributes,describingClothingAttributes,relativeAttributes,
describingObjectsAttributes,sharingFeaturesObjectsAttributes,TIP_attribute1,TIP_attribute2,TIP_attribute3}.
Farhadi et al. \cite{describingObjectsAttributes} learn robust attribute classifier for object description by selecting informative features.
For accurate attribute prediction, Jayaraman et al. \cite{decorrelatingAttributes} propose a method to encourage information sharing among the closely related attributes using structured $ l_{21} $ sparsity regularization. Hwang et al. \cite{sharingFeaturesObjectsAttributes} propose a method for feature sharing between object recognition and attribute perdition.
To learn the semantic attributes for clothing description, Chen et al. \cite{describingClothingAttributes} first extract low-level features based on the pose estimation result, then further improves the attribute prediction accuracy by modeling the mutual dependencies of the attributes with conditional random field.
Instead of predicting the presence of an attribute, Parikh and Grauman \cite{relativeAttributes} propose relative attribute to model the strength of an attribute in an image with respect to other images.
All of the previous methods learn attribute classifier based on low-level features \cite{decorrelatingAttributes,describingClothingAttributes,relativeAttributes,
describingObjectsAttributes,sharingFeaturesObjectsAttributes}. 

Differently, we use deep features extracted by CNN models to learn attributes: both nameable and hidden (data-driven discoverable). 
The deep structure of CNN can produce robust feature generalization ability in different computer vision tasks \cite{imageNetCNN, decafCNN, veryDeepCNNs, representationLearning,shuai2015integrating}.
In attribute learning, we usually don't have enough training data. To tackle this problem, some vision papers use multi-task learning to effectively share visual knowledge \cite{decorrelatingAttributes, sharingFeaturesObjectsAttributes,learningToShareLatent,WangLi-multitask}. Different from their works, we integrate multi-task learning with the powerful CNN models to learn semantic attributes. We jointly learn feature representations and visual knowledge sharing at the same time. We use an enhanced multi-task sharing strategy, where different attribute categories can adaptively select to share from intra-group members, and compete inter-group members, through a flexible decomposition of a shared hidden latent task layer and linear combination layers. In addition, we unsupervisedly train CNN models to discover data-diven attributes. These attributes are important to provide complementary informations beside semantic ones.

Based on these learned representations, we exploit the nature of SPN to model the high-order correlations between attribute representations. SPN is first introduced in \cite{Poon_Domingos_SUM_PRODUCT_ORIGINAL} as a deep probabilistic model, that can capture the deep relationships. The theoretical works \cite{gens2013learning,delalleau:shallow_SPN,rooshenas2014learningDirect} motivate the research of SPN. Delalleau and Bengio \cite{delalleau:shallow_SPN} prove that deep SPN is very efficiently in representing some functions. Gens and Domingos \cite{gens2013learning} propose an algorithm for SPN structure learning. Rooshenas and Lowd \cite{rooshenas2014learningDirect} propose another method for learning SPN structures that can capture both indirect and direct interactions between variables. SPN has been successfully applied in computer vision tasks, including image classification \cite{gens2012discriminative}, image completing \cite{Poon_Domingos_SUM_PRODUCT_ORIGINAL}, and facial attribute analysis \cite{Wang_xiaogang_SPNSforFacialAttributes}.  For the first time, we extend SPN to be a ranking machine by modeling the high-order correlations of its leaf nodes. We also propose a method to learn the parameters of SPN based on pairs of images.

Our work is related to a number of existing high-level image understanding papers, including fashion modeling \cite{chen2013modeling}, interestingness prediction \cite{predictingInterestingnessAesthetics}, image importance prediction \cite{facesAttarctInstgram} and image memorability \cite{understandingMemorability}. But different from them, we are focusing on ranking fashion images, which is considered as an important task for on-line shopping and social websites.

\section{Datasets}
\label{Sec:Dataset}

\textit{Pinterest.com} is one of the most popular websites for image sharing. Users can browse the latest images which are shared by others, and pin the images they like. The number of \textit{likes} can tells the likeabilitiy of an image. It is reported that 80\% of the \textit{Pinterest} users are female \cite{WOMEN_80Percent}. Thus, it is reasonable to collect images from this website to study the likeability of dress images. We collect images from the category of ``Women's fashion".

\textit{Polyvore.com} is a social commerce website allowing people to discover and shop the latest fashion products. The images belonging to different categories (including dresses, shoes, accessories, etc) are uploaded by different online shops.  In this website, users press the button \textit{Like} to show whether they like an image or not. Our technique will be useful for the shops in the website, if they want to identify images which have potential to be liked by many people. They can put up these images in a conspicuous place to attract customers.

We assess the likeability of images based on their number of \textit{likes}. In general, more exposure usually leads to more likes. But on these websites, images with more exposure are usually uploaded by influential/fashion people, who influence many others on what would be liked. This implicitly means such images are likely to be liked more, even based on the same number
of views.

We collect dress images from the above two websites: $ 6,673 $ from \textit{pinterest.com} and $ 69,256 $ from \textit{polyvore.com}. Fashion depends on time. A dress image, which was popular last year, may be not so attractive now. We only collect images which were uploaded in the same month. We track these images and obtain their number of \textit{likes} two months after uploading. Once we use this method for real world application, we also need to continuously update the dataset to learn the new trends. And we only consider high quality images in our dataset, in order to avoid the noise from the image quality itself.

We segment dresses from the images based on the techniques of human detection \cite{lsvm-pami}, face detection \cite{Face_detection_Viola01robustreal-time}, and Grabcut \cite{Grabcut}. As our images are relatively clean (mainly fashion product images) and well posed, these techniques are sufficient to segment dresses from the image.

\section{Attribute Learning}
\label{sec:journalAttributeLearning}

Image representation is critical to predict the likeability of dress images.  Humans describe and judge fashion images based on attributes. 
In this section, we propose to learn semantic attributes and data-driven attributes as middle level representations of dress images. A semantic attribute  represents a certain nameable property of the dress. A data-driven attribute  represents a visual pattern that does not have a name.
To learn semantic attributes, we use a structure of CNN for multi-task learning to share visual knowledge between attributes. To learn data-driven attributes, we propose an unsupervised method based on adapted CNN to discover visual clusters.

\subsection{Semantic Attribute Learning}

Since our datasets have no semantic attribute annotations, we learn a set of attribute classifiers from the clothing attribute dataset of \cite{describingClothingAttributes}, and apply these learned classifiers on our dataset for attribute prediction. We train a binary CNN model for each attribute.

%\begin{figure*}
%    \begin{center}
%        %\fbox{\rule{0pt}{2in} \rule{.9\linewidth}{0pt}}
%        \includegraphics[width=0.75\textwidth]{Figures//newFigure.jpg}
%    \end{center}
%    \caption{Multi-task CNN models: the input image (in the left) with attribute labels information is fed into the model. Each CNN will predict one binary attribute. The shared layer $L$ together with $S$ layer form a weight matrix of the last fully-connected layer followed by soft-max. $L$ layer is a shared latent matrix between all CNN models. Each row in $S$ is CNN-specific weight matrix layer. The soft-max and loss layers are replaced by our multi-task squared hinge loss. Group information are utilized during the training of the network, thus attribute classifiers are encouraged to share visual information within groups and compete each other between groups. Each group contains variable number of attributes and is non-overlapped with others.}
%    \label{fig:CNNStructure}
%\end{figure*}

\begin{table}[!t]
\renewcommand{\arraystretch}{1.3}
\caption{Grouping information used in Clothing dataset\cite{describingClothingAttributes}}
\label{tab:clothingGroups}
\centering
     \begin{tabular}{ | l || c | }
            \hline
            \textbf{Group} &\textbf{Attributes}\\ \hline \hline
            Colors & black, blue, brown, cyan, gray, green, \\
            & many, red, purple, white, yellow \\ \hline
            Patterns & floral, graphics, plaid, solid, stripe, spot \\ \hline 
            Cloth-parts & necktie, scarf, placket, collar \\ \hline
            Appearance & skin-exposure, gender \\ \hline 
        \end{tabular}      
\end{table}

We fine-tune the pre-trained CNN model \cite{imageNetCNN} on the clothing dataset \cite{describingClothingAttributes}.
However, each attribute category does not have many training examples to train the classifier. To solve this problem, we applied our multi-task CNN model \cite{TMM-abrar} to enable different attribute classifiers to share visual knowledge and patterns through a shared layer.

%
%\begin{table}[!t]
%    %        \captionsetup{justification=centering}
%   \label{tab:clothingGroups1} 
%    \centering
%         \caption {Grouping information used in Clothing dataset\cite{describingClothingAttributes}}
%        \begin{tabular}{ | l || c | }
%            \hline
%            \textbf{Group} &\textbf{Attributes}\\ \hline \hline
%            Colors & black, blue, brown, cyan, gray, green, \\
%            & many, red, purple, white, yellow \\ \hline
%            Patterns & floral, graphics, plaid, solid, stripe, spot \\ \hline 
%            Cloth-parts & necktie, scarf, placket, collar \\ \hline
%            Appearance & skin-exposure, gender \\ \hline 
%        \end{tabular}      
%    
%    
%\end{table}

Given the CNN models, we aim to learn the matrix $ W $, which is formed by stacking the parameter matrices of softmax layers of each CNN. The key idea behind the model is to decompose this weight matrix  $ W $ into two matrices $ L $ and $ S $, where the latent $ L $ matrix is the shared layer between all CNN models, and $ S $ is a combination matrix, each column corresponds to one CNN classification layer as follows:
\begin{equation}
    \begin{aligned}
        W = LS
    \end{aligned}
\end{equation}
In this way, each CNN shares visual patterns with other CNN models through the latent matrix $ L $, and all CNN models can collaborate together in the training stage. The benefit is that each CNN can leverage the visual knowledge learned from other CNN models.

In addition, attributes are naturally grouped (as shown in table \ref{tab:clothingGroups}). We encode the grouping information by encouraging attributes to share more (compete) if they belong to the same (different) groups.

%
%
%
%
%
%Encoding the attribute group information will encourage intra-group sharing, so that attribute classifiers share some dimensions in the latent task space. Meanwhile, different groups will compete each other as each of them tries to learn a specific portion from the latent dimension space.

%
%
%
%
%Suppose we have $M$ attributes that belong to $G$ groups. The groups are mutually exclusive (an attribute can be shown in only one group). Structured learning methods like group lasso \cite{treeGuidedLasso} are applied in many areas employing grouping information. In our framework, we encourage intra-group feature sharing, and inter-group feature competition through adapting $L_{21}$ regularization term. Thus we apply this on the combination matrix as $\sum_{k=1}^{K}\sum_{g=1}^{G} \|s_{k}^{g}\|_{2}$, where $K$ is the number of latent tasks (latent dimension space), and $G$ is the number of groups where each group contains a certain number of attributes. The notation $s_{k}^{g}$ represents a number of elements of the $ k $th latent task corresponding to the $ g $th group. This will encourage intra-group sharing, so that attribute classifiers share some dimensions in the latent task space. Meanwhile different groups will compete each other as each of them tries to learn a specific portion from the latent dimension space. Additionally, $L_{1}$ norm is applied on the latent matrix $\|L\|_{1}$ \cite{learningToShareLatent, singleTaskLasso}, to learn more localized visual patterns.

The following cost function is to be minimized simultaneously by all CNN models:
\begin{equation}
    %    \captionsetup{justification=centering}
    \label{eq:hingelossW}
    \begin{aligned}
        \min_{L,S} \sum_{m=1}^{M} \sum_{i=1}^{N_{m}}& \frac{1}{2} [max(0, 1-Y_{m}^{i}(Ls^{m})^{T}X_{m}^{i})]^{2}\\
        &+\mu \sum_{k=1}^{K}\sum_{g=1}^{G}\|s_{k}^{g}\|_{2} + \gamma\|L\|_{1} + \lambda\|L\|^{2}_{F}\\
    \end{aligned}
\end{equation}
For the $m$th attribute category, we denote its model parameter as $Ls^{m}$ and the corresponding training data is ${(X_{m}^{i},Y_{m}^{i})}_{i=1}^{N_{m}} \subset R^{d} \times \{-1,+1\}(m=1,2,...,M)$, where $N_{m}$ is the number of the training samples of the $m$th attribute, and $d$ is the total feature dimension space. In our case, $d$ is $4096$. The last term is the Frobeniuse norm to avoid overfitting.

%Given the previous layer image features $X_{m}^{i}$ from all CNN models (the 4096 dimensional feature), this optimization problem is non-convex in \emph{L} and \emph{S}. We employ a block coordinate descent method, fix \emph{L}, optimize \emph{S}; and then fix \emph{S}, optimize \emph{L} until convergence. These two steps are iteratively repeated until equation \ref{eq:hingelossW} is converged. We employ Accelerated Proximal Gradient Descent (APG) as applied in \cite{learningShareVisualMulticlassDetection} to optimize $L$, meanwhile Smoothing Proximal Gradient Descent (SPGD) \cite{Chen_smoothingproximal, treeGuidedLasso, decorrelatingAttributes} is used to optimize $S$. Then given the shared layer \emph{L} parameters, backpropagation is performed to train the bottom layers in all CNN models.
%
%During training, the input images will be fed into all CNN models, where each CNN has its own label set for this input depending on which attribute to predict. In testing, a feature extraction procedure will be performed through the trained CNN models, and then we apply $ L $ and $ S $ projections onto these features to obtain attributes prediction responses.

\subsection{Data-driven Attribute Learning}
\label{Section_data_drivenattributelabel}

Besides the semantic attributes, some fine visual patterns are also important to gain \textit{likes} from people. In some cases, these visual patterns are essential to discriminate one image from the others. We aim to discover these important unnameable visual patterns in an unsupervised way, and call them data-driven attributes in this paper. Recently, unsupervised data-driven attribute discovery attracts much attention \cite{DataDrivenYU1,DataDrivenRastegar2,DataDrivenMahajan3}.

To be a data-driven attribute, the visual pattern shown in a patch should appear in a large number of images. Data-driven attribute discovery is difficult  mainly due the following reasons. Firstly, we don't have any samples for reference in the data-driven attribute learning. Secondly, we don't know whether a certain attribute present in an image or not. Thirdly, if an attribute presents in an image, we don't know its location. To overcome these challenges, we propose an unsupervised method for data-driven attribute discovery.

In this paper, we focus on local visual patterns and partition the images into a number of patches. To discover meaningful visual patterns from data, it is important to develop powerful feature representations. We leverage the CNN model trained on the ImageNet \cite{ILSVRC_2012}. However, this model is not adapted to our fashion data for feature extraction. We propose a new method that jointly discovers data-driven attributes and adapts the CNN model to our data for more powerful feature representations.

%To use deep learning techniques in this task, we need a well-trained model. However, the labels for the patches are not available to train a attribute learning model. I

In order to obtain such a model, we develop a new algorithm, including three steps: patch feature extraction, unsupervised clustering, and fine-tunning. We conduct these steps iteratively until the CNN model fits our task.

Firstly, we generate $ p $ patches (with overlap) for each dress image. The size of patches is proportional to the size of the image. Each patch can capture the appearance of a specific part.
For example, the first patch in the first row usually captures
the appearance of the right shoulder. For each patch, we extract deep features via CNN which are pre-trained using imageNet \textit{(ILSVRC12 challenge \cite{imageNetCNN})}.  We extract the $ 4,096 $ dimensional features of the first fully-connected layer. 

Secondly, we generate tentative cluster labels for the patches by unsupervised clustering. We perform K-means clustering on the deep features of all patches and obtain a set of over-segmented clusters.  These clusters are agglomerated into $ N_c $ centers based on average link \cite{refine_cluster} to capture the spherical structure. The average link between two clusters \textbf{$ C_1 $} and \textbf{$ C_2 $} are calculated as \begin{equation}\label{Cluster_likn} D(C_1,C_2)=\frac{1}{|C_1|.|C_2|}\underset{x\in C_1}{\sum }\underset{y\in C_2}{\sum }d(x,y) \end{equation}  where $ d(x,y) $ measures the distance between $ x $ and $ y $. The closest two over-segmented clusters are merged together until the number of centers reduces to $ N_c $. In this procedure, we drop the clusters which are small and far from the rest. Each of these $ N_c $ centers corresponds to a data-driven attribute.

Thirdly, with the tentative cluster labels and the data, we adapt the CNN model to this visual pattern discovery task via fine tunning. In the fine-tunned model, the soft-max layer has  $ N_c $ nodes, each corresponding to a cluster. We treat patches from one cluster as the positive training data of the corresponding node. In this way, we obtain an adapted CNN model for these clusters without supervision. The adapted CNN model is expected to generate more suitable feature representation for our task.

We repeat the above three steps for a number of iterations and obtain the final reliable data-driven attributes. Some results are shown in Fig \ref{fig:DataDrivenAttributes}.

%Finally we select the most representative cluster centers as the final data-driven attributes. We define a $ N_I $ dimensional vector $ v $ for each cluster center, and $ v_i=1 $ iff at least one patch of the $ i $th image appear in this cluster. Here, $ N_I $ denotes the number of images. We use $ |v_i| $ to assess the representative ability of cluster $ i $. A larger $ |v_i| $ means a more representative cluster.

\section{Ranking Sum-product Networks}
\label{Sec:spatialRankingSPN}

We represent a dress image as an attribute vector, each element corresponding to the activation of an attribute. A single attribute may contribute to the likeability of an image. However, in many more cases, the combination of several attributes or the correlation between them plays the key role to make the dress image attractive or not. In this work, we capture the high-order correlations of the attributes using the deep structure of SPN.

SPN is a newly proposed deep architecture for correlation modeling and inference. For the first time, we propose to learn SPN as a ranking machine. The root value of the proposed SPN is larger if evaluated with an attribute vector of a more likeable image.
The parameters of the SPN are learned based on the difference between pairs of images. In this section, we first introduce the basic idea of SPN, and then propose the learning algorithm of SPN for ranking.

%SPN is a newly proposed deep architecture for correlation modeling and inference. However, the naive SPN cannot model the spatial relationship between the attributes. This section proposes spatial SPN that can capture the spatial relationships. The root value of the spatial SPN is determined not only by the occurrence of the attributes, but also by the spatial relationship between them.
%
%For the first time, we propose to learn spatial SPN as a ranking machine. The root value of the proposed spatial SPN is expected to be larger if evaluated with an attribute vector of a more likeable image.
%The parameters of the SPN are learned based on the evaluations of pairs of images.

\begin{figure*}[t]
\begin{center}
   \includegraphics[width=0.8\linewidth]{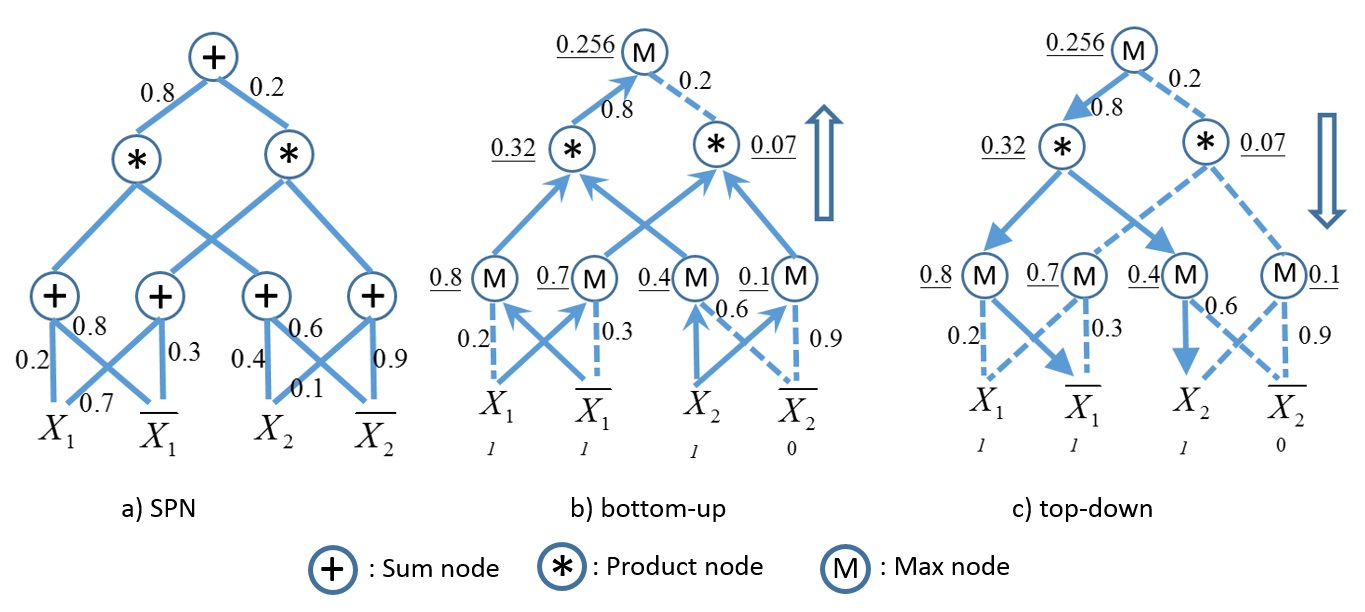}
\end{center}
   \caption{Example of SPN. a) shows a SPN over two variables. b) shows the bottom-up procedure for evaluation with $ x_2=1 $ and marginalized $ x_1 $. c) shows the top-down procedure for inference of the value of $ x_1 $. }
\label{fig:SPNs}
\end{figure*}

\subsection{Sum-Product Networks}

Sum-product Networks (SPN) are directed acyclic graphs with variables as leaves, sums and products of these variables as internal nodes, and weighted edges \cite{Poon_Domingos_SUM_PRODUCT_ORIGINAL}. We introduce the SPN which are built on Boolean variables. Let $x_i$ denote a variable, and $ \bar{x}_i $ denote its negation. 

%The indicator function $ [.] $ is has value $ 1 $ when its argument is true, and 0 otherwise. We abbreviate  the identify indicator $ X_i $ (or $ \bar{X_i} $) by $ x_i $ (or $ \bar{x_i} $). 

The theoretical foundation of SPN is Darwiche's network polynomial \cite{Darwiche:2003:DAI:765568.765570}. Let $ \Phi(x)>0 $ denote an unnormalized probability distribution over a vector of Boolean variables $ x $. The network polynomial of $ \Phi(x) $ is $ \Sigma_x \Phi(x)\Pi(x) $, where $ \Pi(x) $ is the product of indicators that have value 1 in state $ x $. 
%For example, the network polynomial for the joint probability $ \Phi(X_1,X_2) $ of variables $ X_1 $ and $ X_2 $ is $ \Phi(x_1,x_2)x_1x_2+\Phi(x_1,\bar{x}_2)x_1\bar{x}_2+\Phi(\bar{x}_1,x_2)\bar{x}_1x_2+\Phi(\bar{x}_1,\bar{x}_2)\bar{x}_1\bar{x}_2 $. %And the network polynomial for the Bayesian network $ X_1\rightarrow X_2$ is $ P(x_1)P(x_2|x_1)x_1x_2+P(x_1)P(\bar{x}_2|x_1)x_1\bar{x}_2+P(\bar{x}_1)P(x_2|\bar{x}_1)\bar{x}_1x_2+P(\bar{x}_2|\bar{x}_1)\bar{x}_1\bar{x}_2 $.  To calculate the unnormalized probablity of  $ \Phi(x_1,\bar{x}_2) $, we only need to set $ x_1 $ and $ \bar{x}_2 $ to be $ 1 $, and $ \bar{x}_1 $ and $ x_2 $ to be $ 0 $. To calculate the unnormalized probability of $ \Phi(x_1) $, we set the indicator $ x_1 $ to be $ 1 $, $ \bar{x}_1 $ to be $ 0 $, and marginalize the irrelevant variable $ x_2 $, i.e. set both $ x_2 $ and $ \bar{x}_2 $ to be $ 1 $. By setting all indicators to be $ 1 $, we can get the partition function $ Z $. For any evidence $ e $, its probablity can be calculated by $ P(e)=\Phi(e)/Z $. 

With a network polynomial, we can calculate the probability of any evidence easily. However, the size of network polynomial increases exponentially with the number of variables. SPN defined as follows can compactly represent the network polynomial in a hierarchical manner \cite{Poon_Domingos_SUM_PRODUCT_ORIGINAL}.

\textbf{Definition} \cite{Poon_Domingos_SUM_PRODUCT_ORIGINAL} A sum-product network over variables $ x_1,x_2,...,x_d $ is a rooted directed acyclic graph whose leaves are the indicators $ x_1,x_2,...,x_d $ and $ \bar{x}_1,\bar{x}_2,...,\bar{x}_d $ and whose internal nodes are sums and products. Each edge $ (i,j) $ emanating from a sum node $ i $ has a non-negative weight $ w_{ij} $.

%The value of a product node is the product of the values of its children. The value of a sum node is $ \Sigma_{j\in Ch(i)}w_{ij}v_j $, where $ Ch(i) $ are the children of $ i $ and $ v_j $ is the value of node $ j $. The value of a SPN is the value of its root.

Typically, the sum and product nodes are arranged in alternating layers in SPN \cite{Poon_Domingos_SUM_PRODUCT_ORIGINAL,Wang_xiaogang_SPNSforFacialAttributes}. Fig \ref{fig:SPNs} a) shows an example of SPN $ S(x_1,\bar{x}_1,x_2,\bar{x}_2) $ over variables $ x_1 $ and $ x_2 $. Based on this SPN, we can calculate the probability $ P(x_1,\bar{x}_2)=S(1,0,0,1)=0.8(0.2x_1+0.8\bar{x}_1)(0.4x_2+0.6\bar{x}_2)+0.2(0.7x_1+0.3\bar{x}_1)(0.1x_2+0.9\bar{x}_2)=0.8\times0.2\times0.6+0.2\times0.7\times0.9=0.222 $

Using the MPE(Most Probable Explanation) inference \cite{Darwiche:2003:DAI:765568.765570}, SPN can efficiently infer the value of an observed variable. With  the observation of the variable $ x_2=1 $, Fig \ref{fig:SPNs} b) and c) show an example of inference the value of variable $ x_1 $. Firstly, we marginalize the variable $ x_1 $ by setting the input as $ (1,1,1,0) $ and conduct a bottom-up procedure to evaluate the SPN. Then, after replacing the sum nodes with M(maximization) nodes (Fig \ref{fig:SPNs} b), we perform another bottom-up procedure to select the maximum child for each M-node. Finally, we perform a top-down procedure to track the maximum child for each M node and obtain $ x_1=0 $.

\subsection{Learning SPN for fashion image ranking}
\label{sec:learninSPN}

In this subsection, we propose to learn a ranking SPN for image ranking by modeling the relationship of the semantic and data-driven attributes. The deep structure of SPN can model the correlations of any subset of its inputs \cite{Poon_Domingos_SUM_PRODUCT_ORIGINAL}.  Let $ S(I_i) $ denote the evaluation of an SPN with the attribute vector of image $ I_i $ as input. Let $ V(I_i) $ denote the root node value of SPN $ S_i $, which can be considered as unnormalized probability. For a given training set, we generate a set of ordered pairs of images $ P_1=\{(I_h,I_l)\} $, where their difference in number of likes is larger than a threshold $C_1$, i.e. $ n(I_h)-n(I_l)>C_1 $. We also generate another set of un-ordered pairs of images $ P_2=\{(I_a,I_b)\} $, where their difference in number of likes is smaller than a threshold $ C_2$ , i.e. $ |n(I_a)-n(I_b)| \leq C_2 $. 

Our goal is to learn an SPN structure, such that the following two constraints are satisfied: 1) $ \forall (I_h,I_l)\in P_1: V(I_h)>V(I_l) $; and 2) $ \forall (I_a,I_b)\in P_2: V(I_a)=V(I_b) $. In addition, to simplify the structure of SPN, we want to have as few edges as possible. A complicated SPN may overfit the training data, leading to low generalization performance on the testing. Here, we have a constraint for the number of edges $E$, i.e. to be smaller than a threshold $E_0$. A smaller $ E $ means a more simplified SPN structure and a faster speed in inference. We obtain the following objective function
\begin{equation}\label{SPNobjectFunction} 
\begin{split}
 max \qquad & \lambda_1 \underset{(I_h,I_l) \in P_1}{\sum} (V(I_h)-V(I_l)) \\ 
 - & \lambda_2 \underset{(I_a,I_b) \in P_2}{\sum} |V(I_a)-V(I_b)|\\
 s.t. \qquad & E<E_0
\end{split}
\end{equation}
where $ \lambda_1 $ and $ \lambda_2 $ are positive parameters. 

We develop a new method to learn a ranking SPN based on its evaluations on pairs of images. Given an initialized SPN, we first evaluate it with a pair of attribute vectors $ I_1 $ and $ I_2 $ (The pair $(I_1,I_2)$ can be from $P_1$ or $P_2$, the difference will be discussed later). In this procedure, we obtain the value of each node, which can be used in the future inference. To overcome gradient diffusion in SPN training, we convert the SPN $ S $ to an MPN $ M $, by replacing sum nodes with max nodes, as in Fig \ref{fig:SPNs} b). 

Then, aiming to maximize the objective function Eq. \ref{SPNobjectFunction} in log space, we calculate the gradient of log likelihood with regard to the weight $ w $ based on the MPE inference of MPN (to remove the absolute value sign for a pair in $P_2$, we take the image with a larger root value as $I_1$), as follows:
\begin{equation}\label{OurMPNgradient}
\begin{split}
\frac {\partial }{\partial {w}}&log{P}(I_1)-\frac {\partial }{\partial {w}}log{P}(I_2)\\& = \frac{\partial }{\partial _{w}}log\max\Phi(I_1)-
 \frac{\partial }{\partial {w}}log\max\Phi(I_2)
\end{split}
\end{equation}
where $ \Phi(I_1) $ and $ \Phi(I_2) $ represent the maximize polynomial corresponding to the two MPNs.

We use $ M(I_1) $ and $ M(I_2) $ to represent two MPNs, corresponding to $ \max\Phi(I_1) $ and $ \max\Phi(I_2) $.    The partial derivative of the logarithm with respect to the weight $ w_i $ (of the $ i $th edge) as follows          \begin{equation}\label{derivateOfWeight}
\frac{\partial logM(I_1)}{\partial {w_i}}-\frac{\partial logM(I_2)}{\partial {w_i}}=\dfrac{t_i^1}{w_i}-\dfrac{t_i^2}{w_i}
\end{equation}                 where $ t_i^1 $ and  $ t_i^2 $ are the numbers of times that the $ i $th edge is traversed by the MPE inference path in two MPNs. Then, the gradient of the log likelihood of the weight is $ \Delta t_i/w_i $, where $ \Delta t_i=t_i^1-t_i^2 $ is the difference between the number of times that $ w_i $ is traversed when evaluated on the two images.

%\begin{equation}
%F^{HLLC}=\left\{
%\begin{array}{rcl}
%F_L       &      & {0      <      S_L}\\
%F^*_L     &      & {S_L \leq 0 < S_M}\\
%F^*_R     &      & {S_M \leq 0 < S_R}\\
%F_R       &      & {S_R \leq 0}
%\end{array} 
%\right.
%\end{equation} 
%  

%\begin{equation}\label{updateOfWeight} \Delta w_i=\begin{cases}
%
%
%
%\eta\dfrac{\Delta t_i}{w_i} \\
%\eta\dfrac{- \Delta t_i}{w_i}
%
%\end{cases} 
%\end{equation}

For a pair of images $(I_1,I_2) \in P_1$, the learning rate is linearly correlated with the difference between their number of \textit{likes} $\Delta n=(n(I_1)-n(I_2))$, i.e. $ \eta_1 = \alpha_1 \Delta n$.

For a pair of images in $P_2$, we take the image with a larger root value as $ I_1 $ and set the learning rate to be a constant $ \alpha_2 $. Thus, we update the weight $  w_i $ by         \begin{equation}\label{updateOfWeightnew} \Delta w_i=\left\{
\begin{array}{rcl}
\alpha_1 \Delta n \dfrac{\Delta t_i}{w_i}  &  (I_1,I_2)\in P_1\\
\alpha_2 \dfrac{-\Delta t_i}{w_i}   & (I_1,I_2)\in P_2
\end{array} 
\right.
\end{equation}

%Figure \ref{fig:differenceSPN} shows an example of gradient calculation. The inputs of Figure \ref{fig:differenceSPN} a) and b) are respectively the attribute vectors of a more likeable image and a less likeable image. The thick lines trace the top-down procedure of the corresponding MPNs. We take the edge $ e_1 $ as an example to illustrate our optimization method. As the edge $ e_1 $ in Figure \ref{fig:differenceSPN} c) is traversed in $ M(I_h) $ but not in $ M(I_l) $, its weight should be increased. On the contrary, the weight of edge $ e_2 $ should be reduced.

%
%\begin{figure}[t]
%\begin{center}
%   \includegraphics[width=0.98\linewidth]{DifferenceSPN}
%\end{center}
%   \caption{An example of gradient calculation. The first SPN is evaluated with a more likeable image, and the second one is evaluated with a less likeable image. The thick lines in the first two SPNs trace the top-down procedure of the corresponding MPNs. The last SPN shows the gradient of the weights. In the last SPN, the thick line denotes $ \Delta t>0 $ for that edge; the dashed line denotes $ \Delta t<0 $; and the others denote $ \Delta t=0 $.}
%\label{fig:differenceSPN}
%\end{figure}

% % % % % % % % % % % % % % % % % % % % % % % % % % % % %
% % % % % % % % % % % % % % % 

To reduce the number of edges $E$, we cut the unnecessary edges of the SPN in the training procedure. The SPN is updated based on training set according to Eq. \ref{updateOfWeightnew}  for a number of iterations. At the end of an iteration, we investigate the weights of edges that link a sum node and its children. Those edges whose weights are smaller than a threshold are considered as candidates for cutting. For a candidate edge $e_i$, we set its weight to be zero and calculate the object function Eq. \ref{SPNobjectFunction}. We cut this edge $e_i$ if the value of the object function decreases. In this procedure, the nodes with no parent will be deleted to simplify the structure.

To rank a pair of dress images, we take their attribute activation vectors as the input of the learned ranking SPN. The image with more likes is expected to produce a larger root value.

\begin{figure*}[htb]
\begin{center}
%\fbox{\rule{0pt}{2in} \rule{.9\linewidth}{0pt}}
\includegraphics[width=0.8\linewidth]{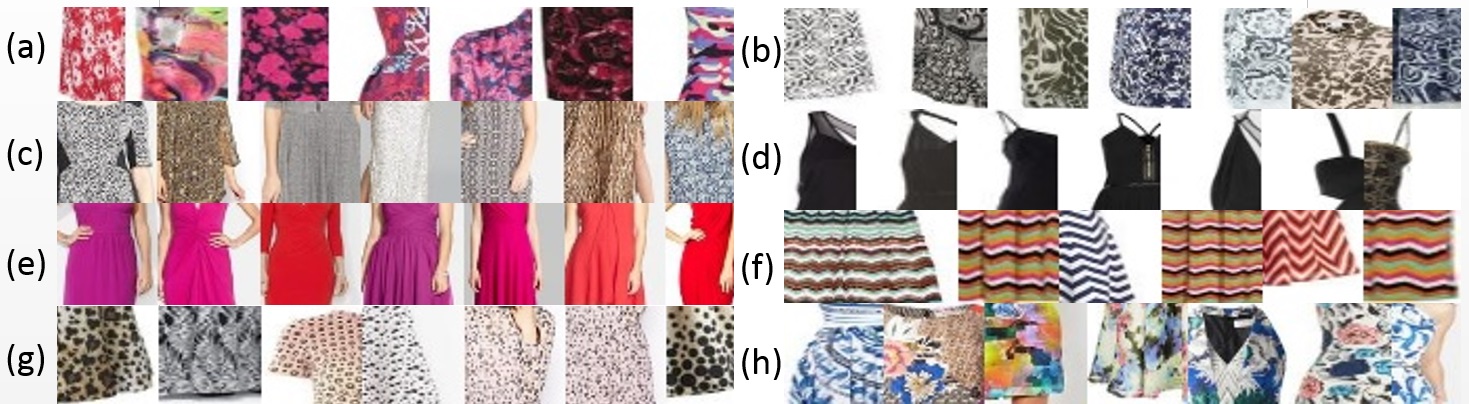}
\end{center}
 \caption{Representative sample patches of eight data-driven attributes. We can observe that each cluster represents a primary visual pattern or color. }
\label{fig:DataDrivenAttributes}
\end{figure*}

\section{Experiments}
\label{Sec:Experiment}

In section \ref{EXP:Seman}, we show the experimental results of multi-task attribute learning on a public clothing dataset. In section \ref{EXP:DataDriven}, we visualize the data-driven attribute learning results. In section \ref{EXP:RANKING}, we learn a ranking SPN for image ranking and compare it with baselines.

\subsection{Semantic attributes}
\label{EXP:Seman}

We evaluate the multi-task learning CNN model on the clothing dataset \cite{describingClothingAttributes}. This dataset contains 1856 images, and 23 binary attributes as shown in Table \ref{tab:clothingGroups}. The nature of these attributes is intended to describe clothing items. The ground-truth is provided on the image-level, each image is annotated to indicate whether it contains a certain attribute.

To train CNN models, we use the code of MatConvNet \cite{VLFEAT} and also used \cite{decafCNN, jia2014caffe}. Our training-testing split per attribute category is half-half.
We compare the multi-task CNN model with four different baselines. 
Baseline 1 (S-CNN) is the traditional single task CNN. Baseline 2 (M-CNN) is a simplified multi-task learning method without the group constraint term in Eq. \ref{eq:hingelossW}. Baseline 3 (ML-CNN) is the CNN with binomial multi-label sigmoid cross-entropy loss. Baseline 4 (G-CNN) is the method of \cite{decorrelatingAttributes} with encoding the group information directly in a quared-hinge loss SVMs. Multi-task CNN with group information (MG-CNN) is the multi-task CNN model \cite{TMM-abrar}, which we used to generate our final image attributes.
Comparing with the previous state-of-the-art results in Conditional Random Field (CRF) \cite{describingClothingAttributes}, the learned CNN models on attributes can outperform all classifiers in \cite{describingClothingAttributes}. After applying multi-task learning, we can further increase CNN model accuracies in almost all cases.

 \begin{table}[h!]
     \caption{Attribute prediction accuracy on the dataset \cite{describingClothingAttributes}. (ML-CNN: binomial multi-label sigmoid cross-entropy loss, CF: combined features model with no pose \cite{describingClothingAttributes}, CRF: method proposed in \cite{describingClothingAttributes}, G-tasks: the method of \cite{decorrelatingAttributes} with encoding the group information directly in a squared-hinge loss SVMs, S-CNN: single task CNN, M-CNN: MTL framework without encoding the group constraint term in Eq. \ref{eq:hingelossW}, and MG-CNN: our whole MTL framework with group constraint. G1: color attributes, G2: pattern group, G3: cloth-parts, G4: appearance group. In G-tasks baseline, the visual features are extracted using our fine-tuned CNN models.)}
     \begin{center}
         \begin{tabular}{ |l||c|c|c|c|c|}
             \hline
             \textbf{Method} &\textbf{G1} &\textbf{G2} & \textbf{G3} &\textbf{G4} &\textbf{Total}\\ \hline \hline
             \textbf{ML-CNN} & 81.97 & 79.11 & 63.16 & 62.45 & 76.25 \\ \hline
             \textbf{CF \cite{compinedFeaturesNoPose}} & 81.00 & 82.08 & 77.63 & 78.50& 80.48 \\ \hline
             \textbf{CRF \cite{describingClothingAttributes}} & 85.00 & 84.33 & 81.25 & 82.50& 83.95 \\ \hline
             \textbf{G-tasks \cite{decorrelatingAttributes}} & 90.10 & 92.29 & 89.80 & 73.52 & 89.18 \\ \hline\hline
             \textbf{S-CNN} & 90.50 & 92.90 & 87.00 & 89.57 & 90.43 \\ \hline
             \textbf{M-CNN} & 91.72 & 94.26 & 87.96 & 91.51 & 91.70 \\ \hline\hline
             \textbf{MG-CNN} & 93.12 & 95.37 & 88.65 & 91.93 & \textbf{92.82} \\ \hline
         
         \end{tabular}
         
     \end{center}
     \label{table:attributePredictionAcc}
  \end{table}

\subsection{Data-driven attributes}
\label{EXP:DataDriven}

We use the data-driven attributes to represent the common visual patterns in the dress images. For this purpose, we propose an unsupervised CNNs adaption procedure in section \ref{Section_data_drivenattributelabel}. The proposed procedure includes three steps in an iterative manner: feature extraction, K-means clustering, and model fine-tunning. The following shows the details of the first iteration.

In our experiments, we randomly select $ 10 $K dress images with relatively high number \textit{likes} from the polyvore dataset. Each image is partitioned into $ 12 $ equally-sized patches with overlap. The size of a patch is proportional to that of the image. This means the sizes of two patches from different images can be different.

We resize these patches and take them as the input of a pre-trained CNNs model \cite{jia2014caffe} using ImageNet \cite{ILSVRC_2012}. The success of  CNNs on large scale image classification means it can extract different deep features for different visual appearance. We use the $ 4,096 $ dimensional feature of the first fully-connected layer to represent a patch.

We hierachically cluster these deep features of $ 120 $K patches. Firstly, we over-segment the feature space into $ 2,000 $ clusters by K-means. Then, we agglomerate these over-segmented clusters into $ 1,000 $ clusters. To capture the spherical structure, this procedure is conducted based the average link. 

We find some clusters only contain patches from a small portion of the images. This means these clusters are not representative enough to be a data-driven attribute. We define a vector $ v^i \in R^{10K} $ to assess the representative ability of the $ i $th cluster. The element $ v_j^i =1 $, if at least one patch of the $ j $th image appear in the $ i $th cluster. Thus, $ \|v^i\|_1 $ counts the number of images that have one or more patches in $ i $th cluster. A larger $ \|v^i\|_1 $ means a more representative cluster. In our experiment, we sort the clusters based on $ \|v^i\|_1 $ and only keep the $ n_c $ most representative clusters, such that

\begin{equation}\label{dataDriven_percentage} 
{\sum_{i=1}^{i= n_c } \|v^i\|_1} \geq 90\% \times {\sum_{i=1}^{i= N_c } \|v^i\|_1}
\end{equation}    
where the number of clusters is reduced from $ N_c $ to $ n_c $. 

With these representative clusters, we fine-tune the CNN model and use it extract features in the next iteration.

We obtain $ n_c=192 $ most representative patch clusters and take them as data-driven attributes. 
Fig \ref{fig:DataDrivenAttributes} shows representative sample patches of $ 8 $ data-driven attributes. We can observe that each data-driven attribute represents a primary visual pattern or color.

Different from semantic attributes, these data-driven attributes represent the local patterns of images. They can be in any location of the image.

\subsection{Image ranking}
\label{EXP:RANKING}

For a dress image, we can use the CNN model trained in section \ref{EXP:Seman} to predict the occurrence of the $ 23 $ semantic attributes and obtain a $ 23 $ dimensional binary vector.

In order to locate the data-driven attributes in an image patch, we use the sliding window method to scan regions. We resize these regions and take them as the inputs of the CNN model obtained in section \ref{EXP:DataDriven}. In this way, we can know the occurrence of the data-driven attributes in an image patch.

We represent each image by a 2327 dimensional attribute activation vector, including $ 192 $ data-driven attributes for each of the $ 12 $ patches, as well as the $ 23 $ semantic attributes. We input such vectors to SPN.

As developing a complete tree structure to link all the possible correlation of these attributes is impossible, we need to learn the structure of SPN for this task. Intuitively, we want to only link the attributes (or attribute sets) which are possibly correlated with each other through edges. Here, we fix the structure of SPN with two stages.

The first stage is to initialize the structure of SPN \cite{Poon_Domingos_SUM_PRODUCT_ORIGINAL}, as following:

Step 1: select a set of subsets of the attributes.

Step 2: for each subset $ R $ , create $ \textit{k} $ sum  nodes $S_1^R,...,S_k^R$, and select a set of ways to decompose $ R $ into  other selected subsets $ R_1,...,R_l $.

Step 3: for each of these decompositions, and for all $ 1\leq i_1,...,i_l \leq k $, create a product node with parents $ S_j^r $ and children $ S_{i_1}^{R_l},...,S_{i_l}^{R_l}  $.\\
In our experiments, the parameter $ k $ equals to $ 10 $. The weights are randomly initialized.

The second stage refines the initialized structure of SPN. We update the parameter to maximize the probability of the samples whose number of \textit{likes} are among the top $ 10 \%$. As neither gradient descent nor EM is  efficient for SPN learning , we update the parameters by hard EM which are used in \cite{gens2012discriminative}. We refine the structure of SPN by removing those children of sum nodes whose weights are zero. 

After fixing the structure, we update the weights of the refined SPN to maximize the difference between the probabilities of image pairs, as proposed in \ref{sec:learninSPN}. The parameter $\alpha_1$ (in eq. \ref{updateOfWeightnew}) is set to be 0.01, and $\alpha_2$ is set to be 0.001. We iteratively update the parameters of SPN for 10 iterations.

To show the effectiveness of representing images using the learned attributes, we compare it with other two baseline representations: low-level hand-designed features and deep features. The low level features include SIFT, LAB color space, LBP, and GIST descriptor feature. The deep feature of a patch is $ 4096 $ dimensional from the first fully connected layer of CNN. We also discover data-driven attributes using \cite{DataDrivenYU1} and \cite{DataDrivenRastegar2}. For classifiers, we take two state-of-the-art machines (ranking SVM (RankSVM \cite{RankSVM}) and structured ranking SVM (S-RankSVM) \cite{rankingstructredSVM}) as baselines. Given a pair of testing images, the ranking is conducted based on the soft responses. Additionally, we also learn SPN without the regularization term of $E$ (number of edges). In this way, the learned SPN is very complicated and turns to be overfitting.

\begin{table}[ht]
\caption{ Image ranking accuracy of RankSVM and Structured RankSVM (LF: low level feature; DF: deep feature; SA: semantic attribute; DA: data-driven attribute; A: semantic attribute and data-driven attribute). }
\begin{center}
    \begin{tabular}{ | l | c | c | c | c| }
    \hline
    \multirow{2}{*}{\textit{Method}} & \multicolumn{2}{c|}{Polyvore} & \multicolumn{2}{c|}{Pinterest}\\
    \cline{2-5}
    & 10 & 20 & 10 & 20 \\
    
   \hline
    \hline
    \textit{RankSVM+LF} & 62.7 &	66.3 &	62.6 &	67.3\\ \hline
    \textit{RankSVM+DF} & 67.4 &	68.8 &	68.5 &	71.2 \\ \hline  
    \textit{RankSVM+SA} & 55.9 &	57.1 &	57.2 &	58.8  \\ \hline
    
   \textit{RankSVM+DA\cite{DataDrivenYU1}} & 65.4 & 66.7 & 	68.4 &	70.3  \\ \hline
     \textit{RankSVM+DA\cite{DataDrivenRastegar2}} & 64.3 & 65.1 & 	67.6 &	69.1  \\ \hline
    
    \textit{RankSVM+DA} & 74.2 & 75.0 & 	74.9 &	75.9  \\ \hline
    \textit{RankSVM+A} & 74.7 &	76.2 &	77.2 &	77.5  \\ \hline
    \hline
        \textit{S-RankSVM+LF} & 62.5 &	66.3 &	63.0 & 66.9	\\ \hline
        \textit{S-RankSVM+DF} & 67.4 &	69.0 &	69.1 &	72.2 \\ \hline  
        \textit{S-RankSVM+SA} & 55.2 & 58.3 &	56.4 & 58.7  \\ \hline
        \textit{S-RankSVM+DA} & 74.8 & 76.2  & 74.9 & 76.4  \\ \hline
        \textit{S-RankSVM+A} & 75.0 & 76.5	 &	77.2 &	77.9  \\ \hline
         \end{tabular}
    \label{tab:SVM_rankresult}
\end{center}
\end{table}

\begin{table}[ht]
\caption{ Image ranking accuracy of Complex SPN and the proposed SPN (LF: low level feature; DF: deep feature; SA: semantic attribute; DA: data-driven attribute; A: semantic attribute and data-driven attribute; C(complex)-SPN: learned SPN without considering the number of edges). }
\begin{center}
    \begin{tabular}{ | l | c | c | c | c| }
    \hline
    \multirow{2}{*}{\textit{Method}} & \multicolumn{2}{c|}{Polyvore} & \multicolumn{2}{c|}{Pinterest}\\
    \cline{2-5}
    & 10 & 20 & 10 & 20 \\
    
   \hline
    \hline
    
    \textit{C-SPN+SA} & 55.4 &	55.8 &	56.3 & 57.4	\\ \hline
    \textit{C-SPN+DA} & 76.2 &	77.3 &	78.0 &	79.3 \\ \hline
    \textit{C-SPN+A} & 77.9 &	78.2 &	78.9 &	80.4 \\ 
    \hline
    \hline
    \textit{SPN+SA} & 58.3 &	58.7 &	57.4 &	60.0\\ \hline

     \textit{SPN+DA\cite{DataDrivenYU1}} & 69.3 &	71.8 &	72.5 &	75.0 \\ \hline
      \textit{SPN+DA\cite{DataDrivenRastegar2}} & 68.4 &	69.2 &	71.4 &	73.6 \\ \hline

        \textit{SPN+DA} & 77.0 &	80.1 &	78.8 &	81.8 \\ \hline
        \textit{SPN+A} & 78.6 &	80.7 &	79.7 &	82.1 \\ \hline
    \end{tabular}
    \label{tab:SPN_rankresult}
\end{center}
\end{table}

The polyvore dataset has $ 12K $ training images and $ 15K $ testing images. The Pinterest dataset has $ 1K $ training images and $ ~3K $ testing images. For testing, we only take the image pairs whose difference number of \textit{likes} is larger than a threshold $ \theta $. Here, we set $ \theta $ to be $ 10 $ or $ 20 $. For the polyvore dataset, we have $ 336K $ testing pairs with $ \theta =10$, and $ 95K $ pairs with $ \theta =20$.
For the Pinterest dataset, we have $ 59K $ testing pairs with $ \theta =10$, and $ 34K $ pairs with $ \theta =20$.
We do not test SPN-based methods on pixel values, low level features, as well as deep features, as our SPN requires binary inputs. In the testing, we consider it is a correct prediction if the root value of the SPN is larger for an image with larger number of \textit{likes}.

Table \ref{tab:SVM_rankresult} shows the ranking accuracies of ranking SVM and structured ranking SVM using different features， when the parameter $ \theta $ is set to be $ 10 $ and $ 20 $. Table \ref{tab:SPN_rankresult} shows the ranking accuracy of SPN-based methods.
For all of these ranking machine, we achieve higher accuracy using mid-level attributes than using low-level features. It indicates that the mid-level attributes are more suitable to our task, as they can tell the visual patterns of the dress images.
We can know from table \ref{tab:SVM_rankresult} and table \ref{tab:SPN_rankresult} that data-driven attributes are better than the semantic attributes in this ranking task.
For example, on Pinterest data, the accuracy of ranking on data-driven attributes is 21\% higher than on semantic attributes.
This is due to two reasons. Firstly, we have many more data-driven attributes than semantic attributes. Secondly, the data-driven attributes are much more discriminative than the semantic attributes. Two images with the same set of semantic attributes can be quite different from each other in appearance.
Compared with ranking SVM and structured ranking SVM, the deep structure of SPN is more powerful to capture the high-order correlations of the attributes, and thus perform better in this ranking task.

\begin{figure}[h]
\begin{center}
   \includegraphics[width=0.9\linewidth]{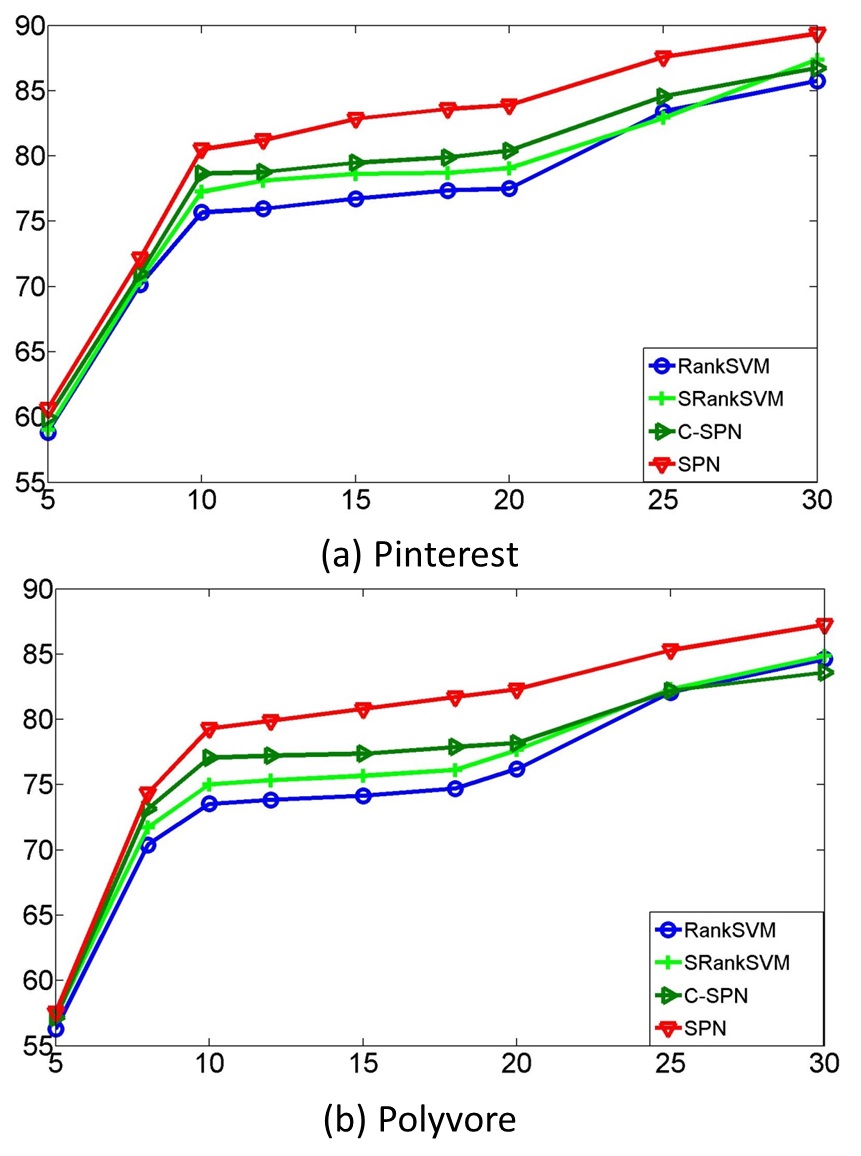}
\end{center}
   \caption{Ranking accuracy (\%) vs. the parameter $ \theta $. In this experiment, we consider image $ I_1 $ is liked by more people than $ I_2 $, if $ I_1 $ has $ \theta $ more \textit{likes} than $ I_2 $. (RankSVM: ranking SVM; SRankSVM: structural ranking SVM; C-SPN: complex SPN; SPN: the proposed method.)}
\label{fig:accuracyVStheta}
\end{figure}

\begin{figure}[h]
\begin{center}
   \includegraphics[width=0.9\linewidth]{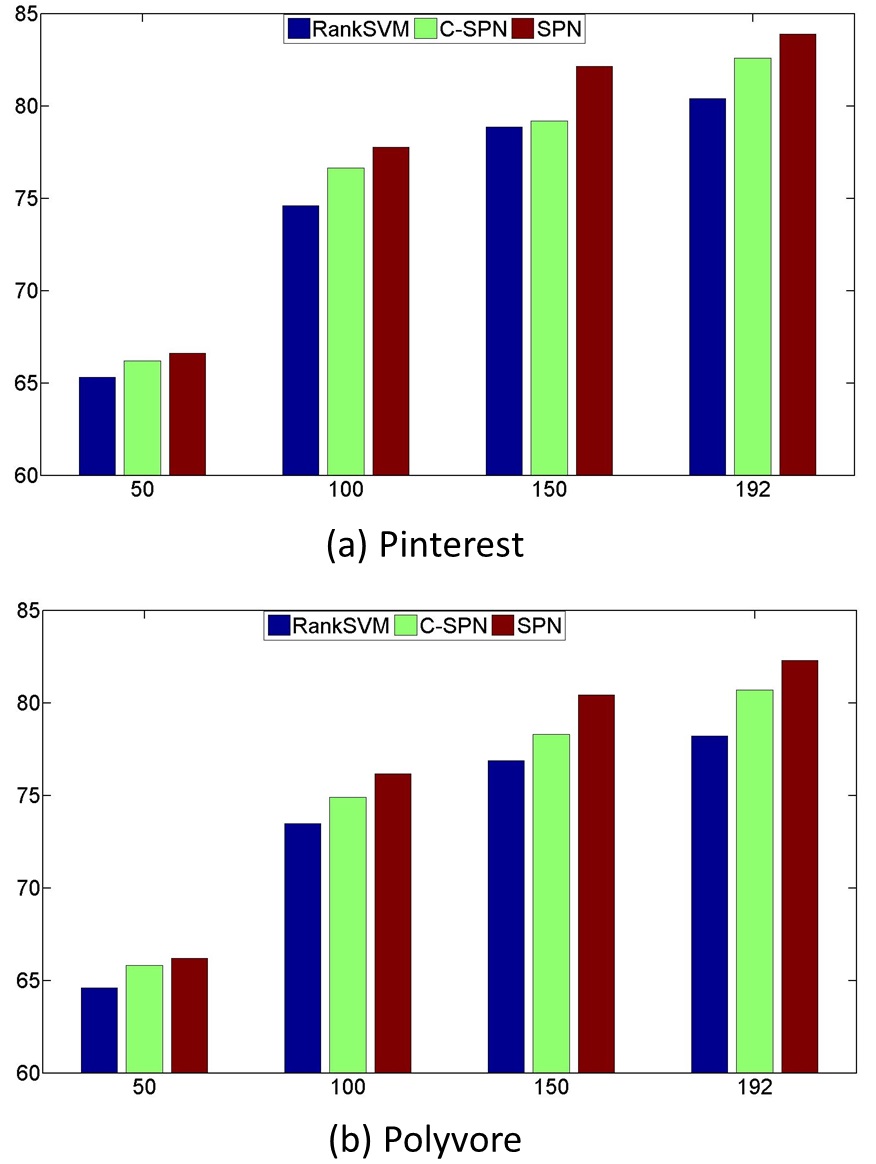}
\end{center}
   \caption{Ranking accuracy (\%) vs. number of data-driven attributes. The horizontal axis denotes the number of data-driven attributes. The vertical axis denotes the ranking accuracy. (RankSVM: ranking SVM; C-SPN: complex SPN; SPN: the proposed method.)}
\label{fig:DAnumber}
\end{figure}

Fig \ref{fig:accuracyVStheta} shows how the ranking accuracy varies with the parameter $\theta$ on our two datasets. A smaller $\theta$ means a lower ranking accuracy mainly due to the following three reasons. Firstly, a smaller $\theta$ means a larger testing set. Secondly, reducing the parameter $\theta$ will make the ranking task more difficult. With a small $\theta$, we have to identity the subtle difference between a pair of images to rank them correctly.
Thirdly, the reliability of ranking order itself reduces as $ \theta $ decreases. For example, it is safe to say an image with $ 100 $ \textit{likes} is liked by more people than an image with $ 2 $ \textit{likes}. However, one image has $ 3 $ more \textit{likes} than another maybe not solely because it is more attractive. Thus, the parameter $ \theta $ should be large enough for real world application.

We conduct image ranking using different number of data-driven attributes. In this experiment, we first sort the data-driven attributes based on their representative ability, which is assessed based on their occurrence probability in the images. Then, we represent the images using semantic attributes and a portion of the most representative data-driven attributes. Fig \ref{fig:DAnumber} shows the ranking accuracies of RankSVM, Complex SPN, and the proposed SPN with different number ($ 50, 100, 150 $ and $ 192 $) of data-driven attributes. With $ 50 $ data-driven attributes, the ranking accuracy of the proposed SPN is only 66.2\% on Polyvore dataset. From $ 50 $ to $ 100 $ data-driven attributes, we can improve the ranking accuracy of SPN by 11.2\%. However, the improvement of ranking accuracy is only 1.76\% from $ 150 $ to $ 192 $ data-driven attributes.

\begin{figure}[h]
\begin{center}
   \includegraphics[width=0.8\linewidth]{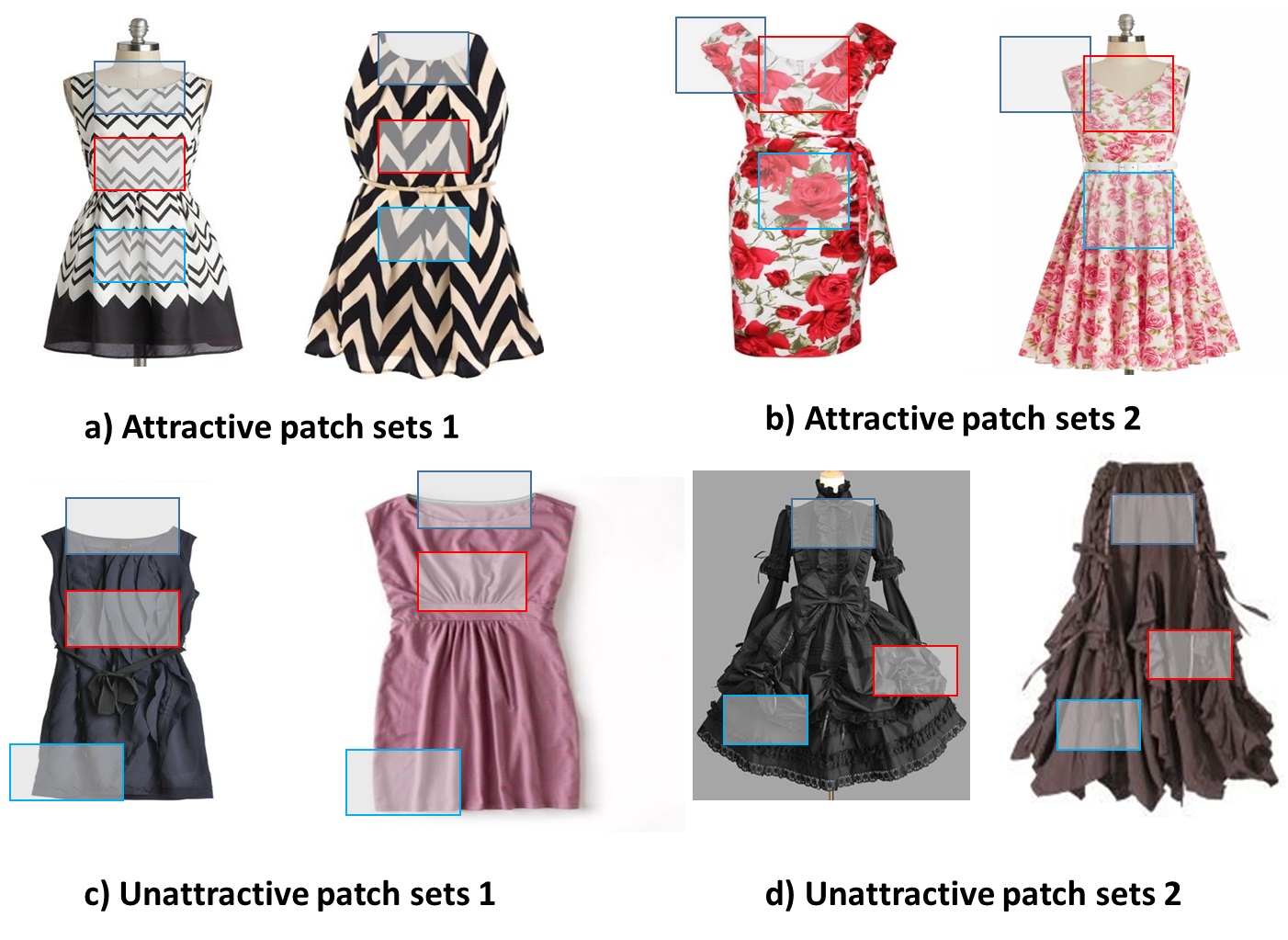}
\end{center}
   \caption{Samples of correlated attractive (a \& b) and unattractive (c \& d) attribute combinations discovered by our SPN model. The rectangles in different colors represent data-driven attributes in different locations.}
\label{fig:coocurence_attribute}
\end{figure}

We also discover attractive and unattractive attribute sets based on learned SPN. To test whether a set of co-occurred attributes can gain more likes or not, we evaluate the SPN with an attribute vector which is only activated in those dimensions corresponding to those attributes. A larger root value means the set can help to raise the likeability of the image. Fig \ref{fig:coocurence_attribute} shows some qualitative samples. The rectangles represent the data-driven attributes in different locations. We can know from fig \ref{fig:coocurence_attribute} b), three colorful data-driven attributes (respectively in shoulder part, neck part, and twist part) co-occur can enhance the likeablity of an image. In contrast, the combinations of the three data-driven attributes shown in c) produce a much lower SPN root value.

\section{Conclusion}
\label{Sec:Conclusion}

This work proposes a method to rank images based on the likeablity of social network community. We normally judge an image based on it visual appearance. Inspired by this, we learn semantic and data-driven attributes as  middle level representation of the images. We capture the high order correlations between these attributes based on a SPN, which can be used to rank images.

%
%\section{Conclusion}
%The conclusion goes here.
%

% or


%% use section* for acknowledgment
%\section*{Acknowledgment}

\ifCLASSOPTIONcaptionsoff
  \newpage
\fi



%\bibliographystyle{IEEEtran}

\bibliographystyle{IEEEtran}
\bibliography{transbib}

\begin{thebibliography}{10}
\providecommand{\url}[1]{#1}
\csname url@samestyle\endcsname
\providecommand{\newblock}{\relax}
\providecommand{\bibinfo}[2]{#2}
\providecommand{\BIBentrySTDinterwordspacing}{\spaceskip=0pt\relax}
\providecommand{\BIBentryALTinterwordstretchfactor}{4}
\providecommand{\BIBentryALTinterwordspacing}{\spaceskip=\fontdimen2\font plus
\BIBentryALTinterwordstretchfactor\fontdimen3\font minus
  \fontdimen4\font\relax}
\providecommand{\BIBforeignlanguage}[2]{{%
\expandafter\ifx\csname l@#1\endcsname\relax
\typeout{** WARNING: IEEEtran.bst: No hyphenation pattern has been}%
\typeout{** loaded for the language `#1'. Using the pattern for}%
\typeout{** the default language instead.}%
\else
\language=\csname l@#1\endcsname
\fi
#2}}
\providecommand{\BIBdecl}{\relax}
\BIBdecl

\bibitem{IncreaseOnlineShopping}
H.~Leggatt, ``Research reveals significant increase in online shopping,'' 2015.

\bibitem{fashionContributeMost}
C.~Baldwin, ``Online spending continues to increase thanks to fashion sector,''
  2014.

\bibitem{Street_to_shop_Yanshuicheng}
S.~Liu, Z.~Song, G.~Liu, C.~Xu, H.~Lu, and S.~Yan, ``Street-to-shop:
  Cross-scenario clothing retrieval via parts alignment and auxiliary set,'' in
  \emph{CVPR}, 2012.

\bibitem{paper_doll_parsing_ICCV2013_berg}
K.~Yamaguchi, M.~Kiapour, and T.~Berg, ``Paper doll parsing: Retrieving similar
  styles to parse clothing items,'' in \emph{ICCV}, 2013.

\bibitem{parsing_clothing_cvpr2012_berg}
K.~Yamaguchi, M.~Kiapour, L.~Ortiz, and T.~Berg, ``Parsing clothing in fashion
  photographs,'' in \emph{CVPR}, 2012.

\bibitem{HipsterWarsECCV14}
M.~H. Kiapour, K.~Yamaguchi, A.~C. Berg, and T.~L. Berg, ``Hipster wars:
  Discovering elements of fashion styles,'' in \emph{ECCV}, 2014.

\bibitem{imageNetCNN}
A.~Krizhevsky, I.~Sutskever, and G.~E. Hinton, ``Imagenet classification with
  deep convolutional neural networks,'' in \emph{NIPS}, 2012.

\bibitem{object_detection_girshick2013rich}
R.~Girshick, J.~Donahue, T.~Darrell, and J.~Malik, ``Rich feature hierarchies
  for accurate object detection and semantic segmentation,'' in \emph{CVPR},
  2013.

\bibitem{ocr_deeplearning}
Y.~Lecun, L.~Bottou, Y.~Bengio, and P.~Haffner, ``Gradient-based learning
  applied to document recognition,'' in \emph{Proceedings of the IEEE}, 1998.

\bibitem{Poon_Domingos_SUM_PRODUCT_ORIGINAL}
H.~Poon and P.~Domingos, ``Sum-product networks: A new deep architecture,'' in
  \emph{ICCV Workshops}, 2011.

\bibitem{decorrelatingAttributes}
D.~Jayaraman, F.~Sha, and K.~Grauman, ``Decorrelating semantic visual
  attributes by resisting the urge to share,'' in \emph{CVPR}, 2014.

\bibitem{describingClothingAttributes}
H.~Chen, A.~Gallagher, and B.~Girod, ``Describing clothing by semantic
  attributes,'' in \emph{ECCV}, 2012.

\bibitem{relativeAttributes}
D.~Parikh and K.~Grauman, ``Relative attributes,'' in \emph{ICCV}, 2011.

\bibitem{describingObjectsAttributes}
A.~Farhadi, I.~Endres, D.~Hoiem, and D.~Forsyth, ``Describing objects by their
  attributes,'' in \emph{CVPR}, 2009.

\bibitem{sharingFeaturesObjectsAttributes}
S.~J. Hwang, F.~Sha, and K.~Grauman, ``Sharing features between objects and
  their attributes,'' in \emph{CVPR}, 2011.

\bibitem{TIP_attribute1}
J.~Shen, G.~Liu, J.~Chen, Y.~Fang, J.~Xie, Y.~Yu, and S.~Yan, ``Unified
  structured learning for simultaneous human pose estimation and garment
  attribute classification,'' \emph{Image Processing, IEEE Transactions on},
  vol.~23, no.~11, pp. 4786--4798, Nov 2014.

\bibitem{TIP_attribute2}
B.~Qian, X.~Wang, N.~Cao, Y.-G. Jiang, and I.~Davidson, ``Learning multiple
  relative attributes with humans in the loop,'' \emph{Image Processing, IEEE
  Transactions on}, vol.~23, no.~12, pp. 5573--5585, Dec 2014.

\bibitem{TIP_attribute3}
Q.~Zhang, L.~Chen, and B.~Li, ``Max-margin multiattribute learning with
  low-rank constraint,'' \emph{Image Processing, IEEE Transactions on},
  vol.~23, no.~7, pp. 2866--2876, July 2014.

\bibitem{decafCNN}
J.~Donahue, Y.~Jia, O.~Vinyals, J.~Hoffman, N.~Zhang, E.~Tzeng, and T.~Darrell,
  ``Decaf: {A} deep convolutional activation feature for generic visual
  recognition,'' in \emph{ICML}, 2013.

\bibitem{veryDeepCNNs}
K.~Simonyan and A.~Zisserman, ``Very deep convolutional networks for
  large-scale image recognition,'' in \emph{ILSVRC}, 2014.

\bibitem{representationLearning}
Y.~Bengio, A.~Courville, and P.~Vincent, ``Representation learning: a review
  and new perspectives,'' in \emph{TPAMI}, 2013.

\bibitem{shuai2015integrating}
B.~Shuai, G.~Wang, Z.~Zuo, B.~Wang, and L.~Zhao, ``Integrating parametric and
  non-parametric models for scene labeling,'' \emph{CRF}, vol.~72, pp. 50--8,
  2015.

\bibitem{learningToShareLatent}
Q.~Zhou, G.~Wang, K.~Jia, and Q.~Zhao, ``Learning to share latent tasks for
  action recognition,'' in \emph{ICCV}, 2013.

\bibitem{WangLi-multitask}
L.~Wang, N.~T. Pham, T.-T. Ng, G.~Wang, K.~L. Chan, and K.~Leman, ``Learning
  deep features for multiple object tracking by using a multi-task learning
  strategy,'' in \emph{Image Processing (ICIP), 2014 IEEE International
  Conference on}, Oct 2014, pp. 838--842.

\bibitem{gens2013learning}
R.~Gens and P.~Domingos, ``Learning the structure of sum-product networks,'' in
  \emph{ICML}, 2013.

\bibitem{delalleau:shallow_SPN}
O.~Delalleau and Y.~Bengio, ``Shallow vs. deep sum-product networks,'' in
  \emph{NIPS}, 2011.

\bibitem{rooshenas2014learningDirect}
A.~Rooshenas and D.~Lowd, ``Learning sum-product networks with direct and
  indirect variable interactions,'' in \emph{ICML}, 2014.

\bibitem{gens2012discriminative}
R.~Gens and P.~Domingos, ``Discriminative learning of sum-product networks,''
  in \emph{NIPS}, 2012.

\bibitem{Wang_xiaogang_SPNSforFacialAttributes}
P.~Luo, X.~Wang, and X.~Tang, ``A deep sum-product architecture for robust
  facial attributes analysis,'' in \emph{ICCV}, 2013.

\bibitem{chen2013modeling}
Q.~Chen, G.~Wang, and C.~L. Tan, ``Modeling fashion,'' in \emph{Multimedia and
  Expo (ICME), 2013 IEEE International Conference on}.\hskip 1em plus 0.5em
  minus 0.4em\relax IEEE, 2013, pp. 1--6.

\bibitem{predictingInterestingnessAesthetics}
S.~Dhar, V.~Ordonez, and T.~L. Berg, ``High level describable attributes for
  predicting aesthetics and interestingness,'' in \emph{CVPR}, 2011.

\bibitem{facesAttarctInstgram}
S.~Bakhshi, D.~A. Shamma, and E.~Gilbert, ``Faces engage us: Photos with faces
  attract more likes and comments on instagram,'' in \emph{SIGCHI}, 2014.

\bibitem{understandingMemorability}
P.~Isola, D.~Parikh, A.~Torralba, and A.~Oliva, ``Understanding the intrinsic
  memorability of images,'' in \emph{NIPS}, 2011.

\bibitem{WOMEN_80Percent}
C.~Smith, ``By the numbers: 140 amazing pinterest statistics,'' last Updated:
  May 24, 2013.

\bibitem{lsvm-pami}
P.~F. Felzenszwalb, R.~B. Girshick, D.~McAllester, and D.~Ramanan, ``Object
  detection with discriminatively trained part based models,'' in \emph{TPAMI},
  2010.

\bibitem{Face_detection_Viola01robustreal-time}
P.~Viola and M.~Jones, ``Robust real-time object detection,'' in \emph{IJCV},
  2001.

\bibitem{Grabcut}
C.~Rother, V.~Kolmogorov, and A.~Blake, ``"grabcut": Interactive foreground
  extraction using iterated graph cuts,'' in \emph{SIGGRAPH}, 2004.

\bibitem{TMM-abrar}
G.~W. Abrar H.~Abdulnabi and J.~Lu, ``Multi-task cnn model for attribute
  prediction,'' in \emph{IEEE TMM}, 2015.

\bibitem{DataDrivenYU1}
F.~Yu, L.~Cao, R.~Feris, J.~Smith, and S.-F. Chang, ``Designing category-level
  attributes for discriminative visual recognition,'' in \emph{CVPR}, 2013.

\bibitem{DataDrivenRastegar2}
M.~Rastegari, A.~Farhadi, and D.~Forsyth, ``Attribute discovery via predictable
  discriminative binary codes,'' in \emph{ECCV12}.

\bibitem{DataDrivenMahajan3}
D.~Mahajan, S.~Sellamanickam, and V.~Nair, ``A joint learning framework for
  attribute models and object descriptions.''\hskip 1em plus 0.5em minus
  0.4em\relax ICCV, 2011.

\bibitem{ILSVRC_2012}
J.~Deng, W.~Dong, R.Socher, L.~Li, K.~Li, and L.~Fei-fei,
  ``\textsc{ILSVRC-2012},'' in
  \emph{http://www.image-net.org/challenges/LSVRC/2012/}, 2012.

\bibitem{refine_cluster}
R.~R. Sokal and C.~D. Michener, ``{A statistical method for evaluating
  systematic relationships},'' in \emph{University of Kansas Scientific
  Bulletin}, 1958.

\bibitem{Darwiche:2003:DAI:765568.765570}
A.~Darwiche, ``A differential approach to inference in bayesian networks,''
  \emph{J. ACM}, vol.~50, no.~3, pp. 280--305, 2003.

\bibitem{VLFEAT}
A.~Vedaldi and B.~Fulkerson, ``Vlfeat: An open and portable library of computer
  vision algorithms,'' 2008.

\bibitem{jia2014caffe}
Y.~Jia, E.~Shelhamer, J.~Donahue, S.~Karayev, J.~Long, R.~Girshick,
  S.~Guadarrama, and T.~Darrell, ``Caffe: Convolutional architecture for fast
  feature embedding,'' \emph{arXiv preprint arXiv:1408.5093}, 2014.

\bibitem{compinedFeaturesNoPose}
A.~Gallagher and T.~Chen, ``Clothing cosegmentation for recognizing people,''
  in \emph{CVPR}, 2008.

\bibitem{RankSVM}
R.~Herbrich, T.~Graepel, and K.~Obermayer, ``Large margin rank boundaries for
  ordinal regression,'' in \emph{Advances in Large Margin Classifiers}, 2000.

\bibitem{rankingstructredSVM}
A.~Mittal, M.~B. Blaschko, A.~Zisserman, and P.~H.~S. Torr, ``Taxonomic
  multi-class prediction and person layout using efficient structured
  ranking,'' in \emph{ECCV}, 2012.

\end{thebibliography}

\end{document}